\pdfoutput=1

\documentclass[11pt]{article}

\usepackage{ACL2023}

\usepackage{times}
\usepackage{latexsym}

\usepackage[T1]{fontenc}

\usepackage[utf8]{inputenc}

\usepackage{microtype}

\usepackage{inconsolata}

\usepackage{xcolor}

\newcommand{\q}[1]{``{#1}''}
\newcommand{\qi}[1]{\textit{``{#1}''}}

\title{The Open-domain Paradox for Chatbots: \\Common Ground as the Basis for Human-like Dialogue}

\author{Gabriel Skantze \\
  KTH Speech, Music and Hearing  \\
  Stockholm, Sweden \\
  \texttt{skantze@kth.se} \\\And
  A. Seza Doğruöz  \\
  Universiteit Gent \\
  Belgium  \\
  \texttt{as.dogruoz@ugent.be} \\}

\begin{document}
\maketitle
\begin{abstract}
There is a surge in interest in the development of open-domain chatbots, driven by the recent advancements of large language models. The \q{openness} of the dialogue is expected to be  maximized by providing minimal information to the users about the common ground they can expect, including the presumed joint activity. However, evidence suggests that the effect is the opposite. Asking users to \q{just chat about anything} results in a very narrow form of dialogue, which we refer to as the \textit{open-domain paradox}. 
In this position paper, we explain this paradox through the theory of common ground as the basis for human-like communication. Furthermore, we question the assumptions behind open-domain chatbots and identify paths forward for enabling common ground in human-computer dialogue. 
\end{abstract}

\section{Introduction}


Recent advancements of large language models (LLMs) have given rise to a surge in interest for the development of \q{open-domain} chatbots \cite{roller_open-domain_2020,Adiwardana2020,thoppilan2022lamda}. Unlike task-oriented dialogue systems designed for a specific purpose and typically implemented in a modular fashion, open-domain chatbots are trained end-to-end on large amounts of  data. 
\citet{roller_open-domain_2020} define their long-term goal as \q{building a superhuman open-domain conversational agent} that is \q{preferred on average to an alternative human speaking partner in open conversation}. The more specific purpose of such a conversational agent is not stated, and thus it is implied that dialogue is a generic problem that can be abstracted away from the context in which it takes place. 


In current evaluations of open-domain chatbots, there seems to be a general assumption that the \q{openness} of the dialogues can be maximized by removing as much instructions and context as possible (e.g., by instructing the user to \q{just chat about anything}).
While this might seem intuitive at first in terms of removing the boundaries for \q{openness}, we argue that this assumption stems from a misconception and that dialogue as a linguistic activity cannot be stripped from its context. The setting in which open-domain chatbots are evaluated does not clearly correspond to any form of human-human dialogue \q{in the wild}. 

In this position paper, we analyse this misconception as the \textbf{open-domain paradox}: \textit{The diversity of the various forms of dialogues found in human-human interaction does not stem from the \q{openness} of the dialogue setting, but rather the opposite: they stem from the diversity of highly specific contexts in which dialogue takes place.}
If this is true, it means that the current methods for collecting dialogue data and evaluating open-domain chatbots will only give rise to a very narrow form of dialogue which does not correspond closely to human-human dialogues. Thus, they will not tell us whether these systems are truly \q{open}. Nor will they tell us much about how good these systems actually are at modelling various dialogue phenomena. From the user's perspective, if the common ground and the reason for having the interaction is not clear, there is a risk that the system will not be perceived as meaningful. 

Our contribution has the following goals: First, we provide a critical review (not an extensive survey) of SOTA open-domain chatbots, in terms of how  they are defined, trained and evaluated. We discuss how the lack of common ground has  consequences for their limited scope and arguably their \q{openness}, compared to human-human dialogue. Secondly, we provide various research directions which might help to mitigate this problem and enable common ground in human-computer dialogue. 


\section{What is common ground?}

When we initiate a dialogue as humans, we do not start with a blank slate but we assume some \textbf{common ground} between the speakers/interlocutors. \newcite{clark1996herbert} describes common ground among humans as \q{the sum of their mutual, common or joint knowledge, beliefs and suppositions}. For a successful and meaningful communication to take place, and for coordinating joint actions, it is essential that both parties have a shared understanding of what this common ground is. 

\newcite{clark1996herbert} makes a distinction between \textit{communal} and \textit{personal} common ground. Communal common ground refers to the cultural communities (e.g., nationality, profession, hobbies, language, religion, politics) people belong to. In addition, there could also be cultural communities which are shaped around shared expertise specific to the members of that community who may not live in the same place (e.g., English teachers around the world), and it is possible to belong to more than one cultural community at the same time. \newcite{clark1996herbert} makes a further distinction in communal common ground between \textit{human nature} (i.e., same senses, sense organs, types of sensations), \textit{communal lexicons} (e.g., there are conventions about word meanings even when two interlocutors speak the same language), as well as \textit{cultural facts, norms and procedures} (which are commonly shared within that community). Procedures for joint activities are the underlying notions about common ground for the community members who know the specific \q{scripts} for the procedures about joint activities in certain contexts (e.g., restaurants, supermarkets vs. school). 

Personal common ground is based on personal joint experiences with someone \cite{clark1996herbert}, and is further classified into \textit{perceptual bases}, \textit{actional bases} and \textit{personal diary}. 
One important aspect of the personal common ground, apart from shared memories and commitments, is the \textbf{linguistic alignment} whereby human interlocutors align (or adjust) their language in alliance with their conversational partners \cite{pickering2006alignment}, context and medium of communication \cite{werry1996linguistic,nguyen2016computational}. Since childhood, humans learn how to adjust and tolerate linguistic variation (e.g., across conversational partners, contexts, mediums) between different communal and personal common grounds in their environment.

The theory of common ground also postulates the principle of \textit{least collaborative effort}, which means that people in conversation use their assumed common ground to minimize their collaborative effort to achieve further understanding \cite{clark1996herbert}. Thus, a brief word might have a significant meaning if the context is highly specific or the interlocutors know each other well. 
As another example, 
\citet{meylan2022adults} showed that conversations between children and their caregivers are hard to transcribe, since their common ground is not known to the transcriber.

An important part of common  ground is the \textbf{joint activity} that is assumed, that is, the reason why the interaction is taking place. This is similar to Wittgenstein's (\citeyear{Wittgenstein1958}) concept of \textit{language games} or the notion of \textit{activity type} developed by \citet{Levinson1979}.
A related concept is that of \textbf{speech events} developed by \citet{goldsmith1996constituting} (not to be confused with \q{speech acts}), which refers to the type of activity that the parties are involved in. By analysing transcribed speech diaries from 48 university students over a 1-week period, they developed a taxonomy of 39 speech events:

\begin{itemize}
    \item \textbf{Informal/Superficial talk}: Small talk, Current events talk, Gossip, Joking around, Catching up, Recapping the day's events, Getting to know someone, Sports talk, Morning talk, Bedtime talk, Reminiscing
    \item \textbf{Involving talk}: Making up, Love talk, Relationship talk, Conflict, Serious conversation, Talking about problems, Breaking bad news, Complaining
    \item \textbf{Goal-directed talk}: Group discussion, Persuading conversation, Decision-making conversation, Giving and getting instructions, Class information talk, Lecture, Interrogation, Making plans, Asking a favor, Asking out
\end{itemize}

This specific taxonomy is likely not generic for all human-human conversations (i.e., there may be more events which they did not identify in their limited study in terms of duration, population and methodology). Nevertheless, it illustrates the diversity of joint activities in human-human interaction. Note that even when we engage in more casual speech events, such as \textit{Small talk}, there is still a reason for why we are having the interaction (maybe just to pass time or avoid being rude), and both interlocutors should be aware of this reason (it is part of their common ground). The speech event or joint activity that is assumed puts constraints on the interpretation space (e.g., which implications can be made) and what can be considered to be a coherent and meaningful contribution to the activity. Thus, in \textit{Small talk} or during \textit{Decision-making}, we expect the speakers to bring up certain topics, but not others. We also do not engage in any speech event with anyone at any time.

\section{Open-domain chatbots}

\subsection{What does open-domain mean?}

The term \q{chatbot} (and its predecessor \q{chatterbot}) has been used since the early 1990’s to denote systems that interact with users in the form of a written chat, typically without any constraints (at least as presented to the user) on what the conversation should be about \citep{mauldin1994chatterbots, wallace2009anatomy}. This early line of work was primarily done outside of academia, where the focus instead was on more task-oriented systems. A search in the DBLP bibliographic database for computer sciences\footnote{http://dblp.org} reveals that the word \q{chatbot} was used in the titles of very few publications until 2015. After this, the usage of the term has increased rapidly, and in 2022, it was used in the titles of almost 300 papers. This development has clearly been sparked by the development of LLMs and the end-to-end modelling of dialogue \citep{Vinyals2015}, which has attracted people from the machine learning community. To stress the open-ended nature of these chatbots, the term \q{open-domain chatbot} is often used. \citet{Adiwardana2020} provide the following definition: \q{Unlike closed-domain chatbots, which respond to keywords or intents to accomplish specific tasks, open-domain chatbots can engage in conversation on any topic}.

Many of the early chatbots were developed to take part in the Loebner prize competition that was running between 1990-2019 \cite{mauldin1994chatterbots}. This competition was partly inspired by the so-called \textit{Turing test}, as proposed by \citet{turing100428} under the name \textit{the Imitation Game}, as a test for determining whether a computer has reached human-level intelligence. It is interesting to note that, as the game is described by \citet{turing100428}, the testers are not provided with any information that would allow them to assume some form of common ground. It is possible that this original idea by Turing has influenced the concept of open-domain chatbots and how they are evaluated. While they are typically not evaluated according to the original Turing test (i.e., the testers are not supposed to guess whether they are interacting with a computer or human), the context-less setting of the interaction is still similar. 

Although earlier chatbots mainly interacted in written form (due to the limited speech recognition performance), open-domain chatbots are nowadays also sometimes built for spoken interaction. One example of this is the \textbf{Alexa Prize}, which is academic competition sponsored by industry to create an open-domain \q{socialbot} for the Amazon Echo device \citep{ram_conversational_2018}. Users of the device can interact with the socialbot from a randomly selected team by just saying \q{Let's chat} to their device. Since the purpose of the socialbot is similar to that of a typical open-domain chatbot, we also include it in our discussion.

In some work, \q{open-domain} seems to be synonymous with more \q{social} (as opposed to task-oriented) interaction, and such systems have been referred to as \q{social chatbots} \citep{shum2018eliza}. \citet{Deriu2020} make a distinction between  \textit{task-oriented}, \textit{conversational} and \textit{question-answering} chatbots, where \textit{conversational} chatbots \q{display a more unstructured conversation, as their purpose is to have open-domain dialogues with no specific task to solve}. These chatbots are built to \q{emulate social interactions} (ibid.). However, it is not entirely clear how the term \q{social} should be understood in this context, as all conversations are \q{social} in the sense that they are used for interpersonal communication. Using the speech event taxonomy by \citet{goldsmith1996constituting}, mentioned above, the speech events that are closest to this notion are perhaps those belonging to \textit{informal/superficial talk}, whereas more task-oriented chatbots or dialogue systems would rather belong to \textit{goal-directed talk}. However, as their analysis shows, the range of speech events in human communication is much more nuanced than this simple distinction would suggest. 

Another problem with the term \q{open-domain} is that it is unclear what \q{domain} refers to. In one interpretation, it could refer to the joint activity or speech event that the interlocutors are engaged in (e.g., small talk, information seeking, decision-making, negotiation). In another interpretation it could mean a wide range of factual topics (e.g., sports, music, travel, math) that are discussed among the interlocutors. These two notions are to some extent orthogonal. For example, the factual topic of travelling could be discussed in the context of various speech events, such as recapping someone's travel experience (\qi{Tell me about your trip to Paris}), asking for travel advise (\qi{What should I see in Paris?}), or planning a trip (\qi{Let's plan a trip to Paris together}). 

If a chatbot is truly \q{open-domain}, we could perhaps expect it to be able to engage in all combinations of speech events and factual topics that we can expect to find in conversations between humans. However, it is unclear whether this is something we could expect from one and the same agent, since we do not expect this from all human-human encounters in real-life settings. Instead, we are selective about what to talk with who and in which way. For example, we can have a conversation with a travel agent in real-life about the costs, insurances, types of sightseeing associated with a trip to Egypt (factual information) but we do not expect her/him to to have a conversation about making a decision about which souvenir to buy, since that is (probably) beyond her/his work definition. Therefore, even human-human conversations are not that open to cover anything across all contexts.



\subsection{Training of chatbots}
\label{sec:training}

Current open-domain chatbots are typically implemented in an end-to-end fashion as transformer-based LLMs, trained to do next-token prediction on large amounts of text data. For the \textbf{Meena chatbot} \citep{Adiwardana2020}, (unspecified) social media conversations were used as training data. \newcite{roller2020recipes} built the \textbf{Blender} chatbot based on the training data collected from Reddit. More recent chatbots, like \textbf{LaMDA} \citep{thoppilan2022lamda}, have been trained using larger, more general datasets (including both dialogue and other public web documents). Similarly, \textbf{GPT-3} \citep{brown2020} is a general-purpose language model that can be used as a chatbot when prompted in the right way. 

While general language models can be used directly as chatbots, their responses will reflect ordinary language use, which might not always align with the desired output in terms of, for example, truthfulness and toxicity (the so-called \q{alignment-problem}). To address this, \citet{thoppilan2022lamda} fined-tuned LaMDA to optimize human ratings of safety and other qualitative metrics. A more sophisticated approach was taken by \citet{ouyang2022} with their model \textbf{InstructGPT}, which uses so-called \q{reinforcement learning from human feedback} (RLHF), where a model of human raters is used during reinforcement learning to optimize the model towards the desired criteria. 

The RLHF approach was also used when training the chatbot \textbf{ChatGPT}\footnote{https://openai.com/blog/chatgpt/}. This kind of model adaptation is interesting from an \q{open-domain} perspective, since the behavior of the chatbot becomes specific to the instructions given to the human raters. 
In the communication around ChatGPT, there is very little information about what the user can expect in terms of its capabilities or the purpose of the interaction besides the fact that it \q{interacts in a conversational way}. Only when interacting with ChatGPT, it becomes clear that its purpose is to serve as some form of AI assistant or interactive search engine, answering factual questions, as well as assisting in writing text and code. However, it refuses to engage in small talk or give opinions. For example, when asked \q{What is your favorite sport?}, it answers \q{As a language model, I do not have personal preferences or feelings, so I cannot have a favorite sport}. In that respect, ChatGPT should perhaps not be seen as an open-domain chatbot (and it is in fact never advertised with those words). In comparison to other chatbots, we do not know much about the evaluation methods and metrics around ChatGPT, and there is not a publication available explaining them to the wider public.

One problem that was identified early on when training chatbots end-to-end is their lack of coherent responses. When asked about their name or favorite sport twice (with some turns in-between), they could give different responses. One way to address this problem was to give them a \textbf{persona}, which is a description of the character that the chatbot is supposed to represent \citep{Zhang2018,Dinan2020,roller2020recipes}. 
While the persona gives some background for the crowdworker or chatbot that can help to improve their internal consistency, it is typically not communicated to the interlocutor beforehand, so it does not really provide any additional common ground. 

\subsection{Evaluation of chatbots}
 
So far, most of the research on how to evaluate open-domain chatbots have focussed on which \textit{metrics} to use when evaluating them \citep{roller_open-domain_2020,mehri_report_2022}. This includes questions such as whether to use human or automatic measures, what questions to ask to raters, and whether to evaluate dialogues on the turn- or dialogue-level (ibid.). For example, in the Alexa Prize, users were asked at the end of the conversation to rate the interaction on a scale between 1 and 5  \citep{ram_conversational_2018}. 

To evaluate the Meena chatbot, \citet{Adiwardana2020} used the metrics \q{sensibleness} (whether the response makes sense in the given context) and \q{specificity} (whether the response is specific to the context or more of a generic nature, like \q{\textit{I don't know}}). The assessment is done by third party observers (crowdworkers), who read the chats and rate them. They showed that a model with lower perplexity scored higher on those metrics. The Blender chatbot \citep{roller2020recipes} was evaluated using ACUTE-Eval method, where two chats are presented next to each other and a crowdworker assess their \q{engagingness} and \q{humanness}. 

For the LaMDA chatbot, \newcite{thoppilan2022lamda} also assess the \q{groundedness} of responses, which is intended to measure whether the model's output is in accordance with authoritative external sources. This should not be confused with the notion of common ground discussed earlier. They also refer to \q{role consistency} which refers to a metric testing whether the agent is performing its tasks in alignment with what is expected from a similar role in a real-life situation (i.e., consistency with the definition of the agent’s role external to the conversation). 

\section{Lack of common ground in \q{open-domain} dialogue}

While the above-mentioned metrics do say something about the relative merits of chatbots, they do not tell us much about their \q{openness}, or the diversity of the speech events they can engage in. When doing so, more attention should be given to the setting in which the chatbots are evaluated. Since there is no natural setting in which these open-domain chatbots are used, crowd workers are typically recruited to interact with them, either for data collection or for evaluation purposes. Although the crowd workers are typically informed about whether they are interacting with a human or a computer, the setting is similar to that of the Turing test mentioned above, in the sense that no information about the assumed common ground is provided, and they are often asked to initiate the interaction with as few instructions as possible. 

For the Meena chatbot, \q{Conversations start with `Hi!' from the chatbot to mark the beginning of the conversation and crowd workers have no expectation or instructions about domain or topic of the conversation} \cite{Adiwardana2020}. For the LaMDA chatbot, crowd workers were instructed to  \q{Start a conversation with the chatbot by posing a question or typing a statement on any topic you want to talk about} \cite{thoppilan2022lamda}. For the Alexa Prize, the users were not provided with any details on what they could expect. Users were asked (through commercials) to just say \q{Let's chat} to their smart speaker in order to initiate the interaction, but no other instructions were provided \citep{ram_conversational_2018}. 

These forms of generic and minimalistic instructions are perhaps chosen to provide as little bias as possible in terms of what topics will be brought up, and to really stress the \q{open domain} nature of the chatbots. However, given what has been discussed above about the importance of common ground and a shared understanding of what the speech event is supposed to be in human-human dialogues, the setting of open-domain chatbots without any common ground is quite unnatural. There is also no physical context or visual cues that could be used to infer any common ground. 
It is hard to find any similar setting for a human-human conversation. Even if we initiate a small talk with a stranger when waiting for the bus, we both know that this is the type of activity we are engaged in, which will guide us in what might be appropriate to talk about in that context. The equivalent would rather be to be randomly connected to a person without any knowledge about that person or about what the conversation is supposed to be about. Can we expect such a setting to give rise to a wide variety of speech events and topics? If not, how do we know if these systems would be able to handle them?

To address this question, \citet{dogruoz-skantze-2021-open} annotated a subset of the publicly released chats from the Meena chatbot \cite{Adiwardana2020} based on the closest speech event category from \citet{goldsmith1996constituting}. The results showed that almost all of them belonged to the \textit{Small talk} category, indicating that they were indeed very limited in scope in terms of speech events. Interestingly, the same was found when annotating the human-human chats that was used as a reference in the evaluation of the Meena chatbot. Those dialogues had been collected by randomly connecting (Google) employees through a chat based system and asking them to converse about anything, to create a similar setting as that for the human-chatbot interactions. 
This indicates that it was not primarily the users' expected (lack of) agency of the interlocutor that limited the scope of the dialogue, but rather the limited instructions about the context for the interaction.

These findings point to what we have referred to as the \textbf{open-domain paradox}: A completely \q{open} setting for conversation, where it is not possible to assume any form of common ground, does not give rise to an \q{open-domain} dialogue, but rather a very limited form of dialogue in terms of both speech events and factual information. 

Whether the setting for evaluating open-domain chatbots actually gives rise to a diversity of speech events has consequences for our understanding of their capabilities. In their evaluation of the Blender chatbot, \citet{roller2020recipes} reported that it was rated on the same level as the human-human chats taken from the Meena evaluation \citep{Adiwardana2020}. However, in an experiment by \citet{dogruoz-skantze-2021-open}, they also tested whether Blender could handle other speech events than the small talk that would normally take place in an \q{open-domain} type of evaluation. This was done by giving a human tester the task of interacting with Blender on a set of different speech events, such as \textit{Decision-making} or \textit{Making plans}. In this evaluation, the chatbot performed much worse (compared to a human interlocutor). This shows that the setting for the interaction and the instructions provided to the testers influence the outcome of the evaluation. 

\section{Enabling common ground} 
\label{sec:enabling}

While the idea of an \q{open-domain} setting for chatbots is quite pervasive in current research, there are also other trends (and forgotten lessons) pointing towards systems where the context is more specific and where the user can potentially assume some form of common ground. In this section, we will discuss those lines of work, and explore to what extent they could increase the diversity of speech events and open a path towards more human-like (and possibly more meaningful) human-computer dialogue. 

\subsection{Repeated interactions}

A clear limitation for building some form of common ground is that most SOTA chatbots can only handle one-time interactions, which limits the number of relevant speech events that might be relevant. If the interlocutors are allowed to have repeated interactions, they could potentially build common ground together across chat sessions, and more diverse speech events might emerge. One step in this direction was proposed by \newcite{xu2021beyond}, who collected and modelled long term conversations, where the speakers learn about each other's interests over time and also refer/discuss issues from past events. The data was collected over 5 chat sessions (each consisting of 14 utterances) through which the speakers talked about topics expanding over days and weeks in order to build a shared history. 
Due to privacy concerns, the crowdworkers were asked to play one of several different roles. While playing their role, they were also asked to pay attention to the previous interactions with the other speakers. 

Although we might expect the participants in subsequent sessions to start with more common ground, it is not self-evident that the user will continue to be interested in interacting with the chatbot over multiple sessions (if they weren't paid) and that meaningful speech events will arise, given the lack of other forms of common ground and reasons for why these repeated interactions are taking place. \citet{xu2021beyond} do not present any analysis of their data that would help to indicate whether their setting in fact leads to more diversity of speech events. 


\subsection{Constraining the speech event}
\label{sec:constrain}

As we have discussed, the absence of contextual cues does not give rise to a variety of speech events, but rather the opposite. Many speech events, like decision-making, do not naturally arise with open-domain chatbots. An alternative would be to instead implement dialogue systems that target a larger variety of more specific contexts and speech events. Examples of this include negotiation \citep{Traum2003}, persuasion \citep{prakken2006}, and presentations \citep{axelsson2020}. 

One form of more constrained setting is that of \textbf{knowledge-grounded dialogue}, where the agent, or one of the crowdworkers during data collection, has access to an external knowledge source, such as Wikipedia \citep{ghazvininejad_knowledge-grounded_2018,li_knowledge-grounded_2022}. 
As discussed in Section \ref{sec:training} above, chatbots such as ChatGPT, which are restricted in what kind of speech events they willing to engage in, can perhaps also be put into this category. It should be noted though that ChatGPT is still lacking in terms of the common ground the user can assume (e.g., which cultural norms can be expected) and to what extent the user can trust the factual correctness of the answers given. The chatbot also does not 
adjust its answers according to the user's level of knowledge, needs, and preferences.
For example, a human librarian would utilise the presumed common ground and interact with the user (e.g., student)
to find out the purpose of the request (e.g., \q{\textit{Why do you need this information?}}, \q{\textit{Is it for a homework?}}), and the level of the user's knowledge (e.g., \q{\textit{Which grade are you at?}}), to recommend resources that fit with its assumptions about the user. 

Another recent example of a system that implements a specific speech event is \textbf{CICERO}, an agent that can play the game of Diplomacy on a human expert level \citep{meta_diplomacy_2022}. Unlike other games, like Chess or Go, Diplomacy does not only rely on the strategy of how to move pieces on the game board, but also on the verbal interaction between the players, where they need to negotiate, build trust, persuade, and potentially bluff, highlighting the joint activity and common ground clearly. 
This type of speech event is also not very likely to take place with an open-domain chatbot. Thus, a plethora of different speech event-specific dialogue systems will likely give rise to a larger diversity of speech events than what can be expected from one open-domain chatbot. 

\subsection{Situated and embodied interaction}

One limitation with chatbots is the lack of physical embodiment or physical situation from which common ground could be inferred. 
In absence of a shared personal history, common ground can to some extent be inferred from cues like our physical appearance (e.g., age or how we dress) and the language/dialect we use. For example, \citet{lau_i_2001} asked participants to estimate the proportion of other students who would know certain landmarks, which they could do very accurately. The situation in which the interaction takes place can also serve as a cue for humans to establish the common ground with other humans and develop joint actions accordingly. 

If we  present the user with an animated avatar instead of an empty chat prompt, it could perhaps help the user to infer more about their potential common ground \citep{kiesler_fostering_2005,fischer_how_2011}. In case of a robot situated in a physical environment, there should be even more contextual cues. For example, in an analysis of interactions with a robot receptionist, \citet{lee_how_2009} note that \q{it seemed that people assumed the robot would have knowledge about his surroundings [...] or places relevant to his background or occupation}, and thus most questions directed towards the robot were also related to its role and situation.  
Studies have also shown that people use the robot's presumed origin \citep{sau_lai_lee_human_2005} or gender \citep{powers2005} to infer the robot's knowledge (and thereby their common ground).

\subsection{Scenario-based evaluation}

As discussed earlier, \citet{dogruoz-skantze-2021-open} evaluated the Blender chatbot on various speech events by giving the user (tester) specific instructions on which speech event to engage in. For example, for the \textit{Decision-making} speech event, the tester could say to the chatbot: \q{\textit{We have 1000 dollars. Let’s decide how we spend it together}}. By providing the tester with a list of different speech events, it is possible to better understand which of them the chatbot can handle. If further developed, we think this could constitute an interesting evaluation scheme. To increase the common ground, both the chatbot and the user should probably be given a more detailed description of the setting for the interaction. A potential drawback of this evaluation scheme is that it involves crowdworkers who would need to role-play (likely without much engagement in the task), rather than naturally arising speech events, and that there is a limit as to how detailed the scenario descriptions can be. Thus, they would still not be very close to the level of common ground we can expect from human-human dialogue.

\subsection{Simulated worlds}

Multi-player text-adventure games \cite{urbanek-etal-2019-learning} could also provide interlocutors with some common ground. For example, \citet{ammanabrolu_how_2021} present a model for such agents using LLMs and reinforcement learning. For these agents, the context of the game provides common ground in terms of a textual \q{setting}, which describes the reason for why the interaction takes place and of the characters involved in the interaction.

We can imagine even more open worlds (ideally multi-modal) in which agents and/or humans interact with each other. In such settings, a larger variety of speech events (similar to human-human interactions), can be expected, and an agent acting in that world will have to be aware of the context and presumed common ground in order to engage in those speech events. 
An interesting step in this direction was presented in \citet{park_generative_2023}, where a large language model (GPT-4) was used to simulate agents in a virtual world. In this setting, the authors observed social behaviours \q{emerging}. Although they did not use the speech event categorization in their analysis, it is clear from their examples that various speech events took place, including \textit{Small talk}, \textit{Catching up} and \textit{Making plans}; eventually, the agents started to plan a Valentine’s Day party and set up dates with one another. Such simulations could be an interesting setting for studying and modelling diverse forms of dialogue and speech events. While this simulation did not include any human interlocutors, it is easy to see how that could be added. 

\section{Conclusion and Discussion}

Our goal in this paper was not to survey the latest literature on chatbots but to question the assumptions behind the term \q{open-domain}, and scrutinize to what extent chatbots labelled as such are truly \q{open}. We discussed the notion of common ground in human-human dialogue, and how it is important for human-like dialogue and a diversity of speech events and topics. The general assumption behind SOTA open-domain chatbots is instead to remove as much context as possible, often presenting users with an empty prompt and asking them to \q{just chat}. However, both linguistic theory and evidence suggests that the absence of context does not give rise to a diversity of speech events, but rather a very limited form of dialogue. We called this the \textit{open-domain paradox}.

To be able to study and model different forms of dialogue between humans and agents, the dialogue needs to be embedded within a highly specific context, where both the agent and the human can assume some form of common ground. We identified a couple of different paths towards this end in Section~\ref{sec:enabling}.

 

One explanation for the huge interest in the development of (context-less) open-domain chatbots is perhaps that it fits well with the LLM paradigm (next-token prediction), which uses a limited prompt with the dialogue history. In a way, it is an example of a solution that has found its problem. It is of course possible to include a larger context in the prompt (i.e., a textual representation of the common ground that could be expected), but this clearly has its technical limits. It will be interesting to see whether there will be a movement towards other solutions, perhaps using more modular architectures (as in the example of CICERO). 


From the perspective presented here, open-domain chatbots (as the term is currently used) are not necessarily more \q{generic} than task-oriented dialogue systems, given the limited form of dialogue they are typically evaluated against. One option would be to re-brand open-domain chatbots as \q{small talk chatbots} or, as some have suggested, \q{social chatbots} \citep{shum2018eliza}. We do not think this is appropriate either, since even small-talk (between humans at least) is dependent on the presumed common ground between the speakers. We do not exclude the possibility that the type of dialogue crowdworkers have with open-domain chatbots (i.e., dialogue without common ground) can be regarded as a special speech event category, which we have no existing terminology for yet.

For future work, it might be better to characterize dialogue systems based on which contexts they are intended to be used in, and what speech events are expected to take place. When evaluating such systems, it is also important that they are used in the context they were intended for.

As we have discussed, according to the \q{principle of least collaborative effort}, humans use common ground to make their interactions more efficient \citep{clark1996herbert}. Thus, users of dialogue systems will likely always prefer to use systems where their common ground is maximized, rather than \q{open-domain} settings. If one would want to develop a generic \q{superhuman open-domain conversational agent} \citep{roller_open-domain_2020}, it would need to be highly context-aware, in order to serve across contexts. This route is perhaps not very realistic, given the incredible richness and diversity of the forms of common ground  humans assume and build together. Also, this does not even exist for human-human dialogues, as we do not speak about anything with anybody randomly at any given time, without any common ground. Instead of open-domain dialogue systems, it might be more fruitful to focus on developing a large variety of highly context-specific dialogue agents. 

\bibliography{custom}
\bibliographystyle{acl_natbib}




\end{document}